\documentclass[letterpaper]{article} 
\usepackage{aaai23}  
\usepackage{times}  
\usepackage{helvet}  
\usepackage{courier}  
\usepackage[hyphens]{url}  
\usepackage{graphicx} 
\usepackage{natbib}  
\usepackage{caption} 
\usepackage{algorithm}
\usepackage{algorithmic}
\usepackage{caption}
\usepackage{subcaption}
\usepackage{newfloat}
\usepackage{listings}
\DeclareCaptionStyle{ruled}{labelfont=normalfont,labelsep=colon,strut=off} 
\floatstyle{ruled}
\newfloat{listing}{tb}{lst}{}
\floatname{listing}{Listing}
\title{Integrating Policy Summaries with Reward Decomposition for Explaining Reinforcement Learning Agents}
\author{
    Yael Septon, \textsuperscript{\rm 1} 
    Tobias Huber, \textsuperscript{\rm 2} 
    Elisabeth Andr{\'e}, \textsuperscript{\rm 2} 
    Ofra Amir, \textsuperscript{\rm 1} 
}
\affiliations{
    \textsuperscript{\rm 1} Technion, Israel Institute of Technology\\

    \textsuperscript{\rm 2}
    Chair for Human-Centered Artificial Intelligence, University of Augsburg \\
    
    \{yael123, oamir\}@technion.ac.il ,\{tobias.huber, andre\}@informatik.uni-augsburg.de

}

\usepackage{bibentry}

\usepackage{amsfonts}
\usepackage{amsmath}
\usepackage{xcolor}

\begin{document}

\maketitle

\begin{abstract}
Explaining the behavior of reinforcement learning agents operating in sequential decision-making settings is challenging, as their behavior is affected by a dynamic environment and delayed rewards. Methods that help users understand the behavior of such agents can roughly be divided into local explanations that analyze specific decisions of the agents and global explanations that convey the general strategy of the agents. In this work, we study a novel combination of local and global explanations for reinforcement learning agents. Specifically, we combine reward decomposition, a local explanation method that exposes which components of the reward function influenced a specific decision, and HIGHLIGHTS, a global explanation method that shows a summary of the agent's behavior in decisive states. We conducted two user studies to evaluate the integration of these explanation methods and their respective benefits. Our results show significant benefits for both methods. In general, we found that the local reward decomposition was more useful for identifying the agents' priorities. However, when there was only a minor difference between the agents' preferences, then the global information provided by HIGHLIGHTS additionally improved  participants' understanding.  
\end{abstract}

\section{Introduction}
Artificial Intelligence (AI) agents are being deployed in a variety of domains such as self-driving cars, medical care, home assistance, and more. With the advancement of such agents, the need for improving people's understanding of such agents' behavior has become more apparent. In this work, we focus on explaining the behavior of agents that operate in sequential decision-making settings, which are trained in a deep reinforcement learning (RL) framework. This is challenging, as the behavior of RL agents is affected by the dynamics of the environment, the reward specification, and their ability to attribute delayed outcomes to their actions.

We study the effectiveness of providing users with global and local explanations of the behavior of RL agents.
Global explanations explain the general behavior of the agent, e.g., by describing decision rules or strategies. 
In contrast, local explanations try to explain specific decisions that an agent makes. 
While local explanations can provide detailed information about single decisions of an agent, they do not provide any information about its behavior in different contexts.

Prior work investigated a combination of the complementary benefits of global and local explanations.
They compared local saliency maps, that highlight relevant features within the input, with global policy summaries that demonstrate the behavior of agents in a selected set of world states \cite{huber2020local}.
While their results were promising, the local saliency maps were lacking since they were hard for users to interpret correctly.

In this paper, we propose and evaluate a novel combination of global policy summaries with local explanations based on reward decomposition.
Reward decomposition aims to reveal the agent's reasoning in particular situations by decomposing the rewards into reward components, explicitly revealing which reward components the agent expects from each action~\cite{juozapaitis2019explainable}. 
For example, in a driving environment, such reward components could be a reward for driving safely, a reward for driving fast, etc.
Since reward decomposition is more explicit in reflecting the agent's decision-making than saliency maps, we hypothesized that it will be easier to interpret. 
Furthermore, reward decomposition is incorporated directly into the agents' underlying decision model through its training.
In contrast, saliency maps are generated after the agent is trained and might not be faithful to the agents' underlying decision model~\cite{rudin19,huber2022benchmarking}.
Therefore, we hypothesized that a combination of policy summaries with reward decomposition will improve users' understanding of the agents' strategy compared to only using local or global explanations.

We conducted two user studies in which participants were randomly assigned to one of four different conditions that vary in the combination of global and local information: (1) being presented or not presented with a local explanation (reward decomposition), and (2) being presented with a global explanation in the form of a HIGHLIGHTS policy summary~\cite{amir2018highlights} or being presented with frequent states the agent encounters (a baseline for conveying global information).
We used a Highway and a Pacman environment and trained agents that varied in their priorities by modifying the reward function.
Participants were asked to determine the priorities of these agents based on the explanations in their condition.

Our results show that the use of reward decomposition as a local explanation helped users comprehend the agents' preferences.
In addition, the HIGHLIGHTS global explanation helped users understand the agents' preferences in the environment of Pacman.
While we found that the benefit of presenting reward decomposition was greater than that of providing HIGHLIGHTS summaries, the combined explanations further helped users to differ between the agents' priorities when there only was a minor difference between the agents' preferences.

\section{Related Work}
Explainable reinforcement learning methods can broadly be divided into two classes based on their scope: local and global explanations \cite{molnar2022}.
\emph{Local} explanations analyze specific actions of the agent.
This is often further divided into post-hoc and intrinsic explainability methods.
Post-hoc methods analyze the agent after training, for example by creating saliency maps~\cite{hilton2020understanding,huber2019,puri2020}.
Intrinsic methods are built into the agent's underlying decision model to make it more explainable.
For example, in reinforcement learning, this is done for casual explanations ~\cite{madumal2020distal} or by reward decomposition~\cite{juozapaitis2019explainable}.
In this paper, we focus on reward decomposition as a local explanation method.
In reward decomposition, the Q-value is decomposed into several components ~\cite{van2017hybrid}. 
This approach has been used for explainability ~\cite{juozapaitis2019explainable} as the decomposition of the reward function into meaningful reward types can help reveal which rewards an agent expects from different actions. 
A user study exploring the usefulness of different local RL explainability methods showed that reward decomposition contributed to people's understanding of agent behavior~\cite{anderson2019explaining}.

\emph{Global} explanations attempt to describe the high-level policy of an agent, for example by extracting logical rules  that describe the agent's strategy~\cite{booth2019evaluating}. In other work, the goal was to enable users to correctly anticipate a robot’s behavior in novel situations. The key idea was to select states that are optimized to allow the reconstruction of the agent's policy ~\cite {huang2019enabling}.
In this work, we utilize strategy summarization ~\cite{amir2019summarizing} as a global explanation method. Strategy summaries demonstrate an agent's behavior in carefully selected world states. The states can be selected based on different criteria, e.g., state importance~\cite{amir2018highlights} or using machine teaching approaches~\cite{lage2019exploring}.

Most closely related to our work is a study by Huber et al.~\cite{huber2020local} which combined local and global explanation methods in RL agents. They used strategy summaries (global explanation) with saliency maps (local explanation). Since this study showed that using saliency maps as local explanations is lacking, we study the integration of reward decomposition as an alternative local explanation, together with policy summaries. In addition, reward decomposition has the advantage of faithfully representing the underlying decision model, while saliency maps are post-hoc explanations that may not accurately reflect what the model learned~\cite{rudin19,huber2022benchmarking}. 

\section{Background}
\label{sec:background}
We assume a Markov Decision Process (MDP) setting. Formally, an MDP is defined by a tuple $<S, A, R_{a}, Tr>$:
\begin{itemize}
  \item $S$: Set of states.
  \item $A$: Set of actions. 
  \item $R_{s,a,s'}$: The reward received after transitioning from state $s$ to state $s'$, due to action $a$.
  \item $Tr$: A transition probability function $Tr(s,a,s'): S \times A \times S \rightarrow [0,1]$ defining the probability of transitioning to state $s'$ after taking action $a$ in $s$.
\end{itemize}
The Q-function is defined as the expected value of taking action $a$ in state $s$ under a policy $\pi$ throughout an infinite horizon while using a discount factor $\gamma$:
\\ $Q^{\pi}(s,a)= \mathbb{E}^{\pi}[\sum_{t=0}^{\inf}\gamma^{t}R_{t+1}| s_t=s,a_t=a]$.

To learn this Q-function, we used deep Q-Networks (DQN) \cite{mnih2015human}.
A DQN is a multi-layered neural network that for a given state $s$ and action $a$ outputs q-value $Q(s,a;\theta)$, where $\theta$ are the parameters of the network.
During training, the DQN contains two networks, the target network and the value network.
The target network, with parameters $\theta^-$, is the same as the value network except that its parameters are copied every $\tau$ steps from the value network i.e. $\theta^-_t$= $\theta_t$ and kept fixed on all other steps.
The value network is trained by minimizing the sequence of loss functions: $L_t=\mathbb{E}_{s,a,r,s'}[(Y^{DQN}_t-Q(s,a;\theta_t))^2]$, where the target $Y^{DQN}_t$ is given by the target network: $Y^{DQN}_t\equiv R_{t+1}+\gamma \max\limits_{a}Q(S_{t+1},a;\theta^-_t)$.
In this work, we use an improvement of the DQN called Double DQN \cite{van2016deep}.
Double DQN replaces the target $Y^{DQN}_t$ with $Y^{DoubleDQN}_t\equiv R_{t+1}+\gamma Q(S_{t+1}, \operatornamewithlimits{argmax}\limits_a Q(S_{t+1},a;\theta_t), \theta^-_t)$. 
The update to the target network stays unchanged from DQN and remains a periodic copy of the value network.

The exact architecture of the networks we used was specific for each environment and will be described in the corresponding sections.

\subsection{Reward Decomposition}

Van Seijen et al.~\cite{van2017hybrid} proposed the Hierarchical Reward Architecture (HRA) model. 
HRA takes as input a decomposed reward function and learns a separate Q-function for each reward component.
In a game like Pacman, such reward components could for instance correspond to dying or reaching specific goals.
Because each component typically only depends on a subset of all features, the corresponding Q-function can be approximated more easily by a low-dimensional representation, enabling more effective learning.

This can be incorporated in the MDP formulation by specifying a set of reward components $C$ and decomposing the reward function $R$ into $|C|$ reward functions $R_c(s,a,s')$. 
The objective for the HRA agent remains the same as for traditional Q-learning: to optimize the overall reward function $R(s,a,s') = \sum_{c \in C}R_c(s,a,s')$. 
HRA achieves this by training several Q-functions $Q_{c}(s,a)$ that only account for rewards related to their component $c$.
For choosing an action for the next step, the HRA agent uses the sum of these individual Q-functions: $Q_{HRA}(s,a) := \sum_{c \in C} Q_c(s,a)$.
For the update $Y^{DoubleDQN}_t$ each head is used individually.
If the underlying agent uses deep Q-learning, the different Q-functions $Q_{c}(s,a)$ can share multiple lower-level layers of the neural network.
In this case, the collection of Q-functions that each have one type of reward can be viewed as a single agent with multiple \emph{heads}, such that each head calculates the action-values of a current state under his reward function.

HRA was originally proposed to make the learning process more efficient.
However, \citet{juozapaitis2019explainable} suggested the use of \emph{Reward Decomposition (RD)} as a local explanation method.
Traditional Q-values do not give any insight into the positive and negative factors contributing to the agent's decision since the individual reward components are mixed into a single reward scalar.
Showing the individual Q-values $Q_{c}(s,a)$ for each reward component $c$ can explicitly expose the different types of rewards that affect the agent's behavior.

This increase in explainability should not result in a decreased performance.
\citet{van2017hybrid} already showed that HRA can even result in increased performance for Pacman.
We additionally conducted a sanity check in the Highway environment to verify that HRA results in comparable learning to that obtained without decomposing the reward function (see Appendix \ref{ap:sanity_checks}).

\subsection{Policy Summaries}
\emph{Agent strategy summarization} ~\cite{amir2019summarizing} is a paradigm for conveying the global behavior of an agent. In this paradigm, the agent's policy is demonstrated through a carefully selected set of world states. The goal of strategy summarization is to choose the subset of state-action pairs that best describes the agent's policy. In a formal way, Amir \& Amir~\cite{amir2018highlights} defined the set $T= <t_1,...,t_k>$ as the trajectories that are included in the summary, where each trajectory is composed of a sequence of $l$ consecutive states and the actions taken in those states, $<(s_i,a_i),...,(s_{i+l-1},a_{i+l-1})>$. Since it is not feasible for people to review the behavior of an agent in all possible states, $k$ is defined as the size of the summary e.g $|T|=k$.

We use a summarization approach called HIGHLIGHTS ~\cite{amir2018highlights} that extracts the most ``important'' states from execution traces of the agent. 
The importance of a state $s$ is denoted as $I(s)$ and is defined differently between environments as it is influenced by the different actions that are possible in the environment. 
The general idea is that a state is important if the outcome of the chosen action had a big impact. 

\section{Integrating Policy Summaries and Reward Decomposition}
We combined HIGHLIGHTS as a global explanation with reward decomposition as a local explanation. 
We used HIGHLIGHTS to find the most important states during the agents' gameplay. 
For each state that was chosen, we created reward decomposition bars that depict the decomposed Q-values for actions in the chosen state (see Figures \ref{fig:example_from_survey} and \ref{fig:Survey_example_Pacman}).
We chose to combine these two types of explanations because we believe they complement each other.
Reward decomposition reflects the intentions of an agent while HIGHLIGHTS gives a broader perspective on the agent's decisions.

HIGHLIGHTS summaries are typically shown as videos. However, the reward decomposition bars are static, and vary for each state. Therefore, when integrating the two methods, we used HIGHLIGHTS to extract the important states, but displayed them using static images rather than videos.

\section{Empirical Methodology}
To evaluate the benefits of integrating HIGHLIGHTS with reward decomposition as well as their respective contributions to users' understanding of agents' behavior, we conducted two user studies in which participants were asked to evaluate the preferences of different agents. 
We hypothesized that the combined explanations would best support participants' ability to correctly identify agents' preferences and that both the local and global explanations would be better than the baseline information.

\subsection{Experimental Environments and Agent Training}
\label{sec:exp_domains}
\subsubsection{Highway Environment}
\begin{figure}[t]
\centering
\includegraphics[width=0.8\linewidth]{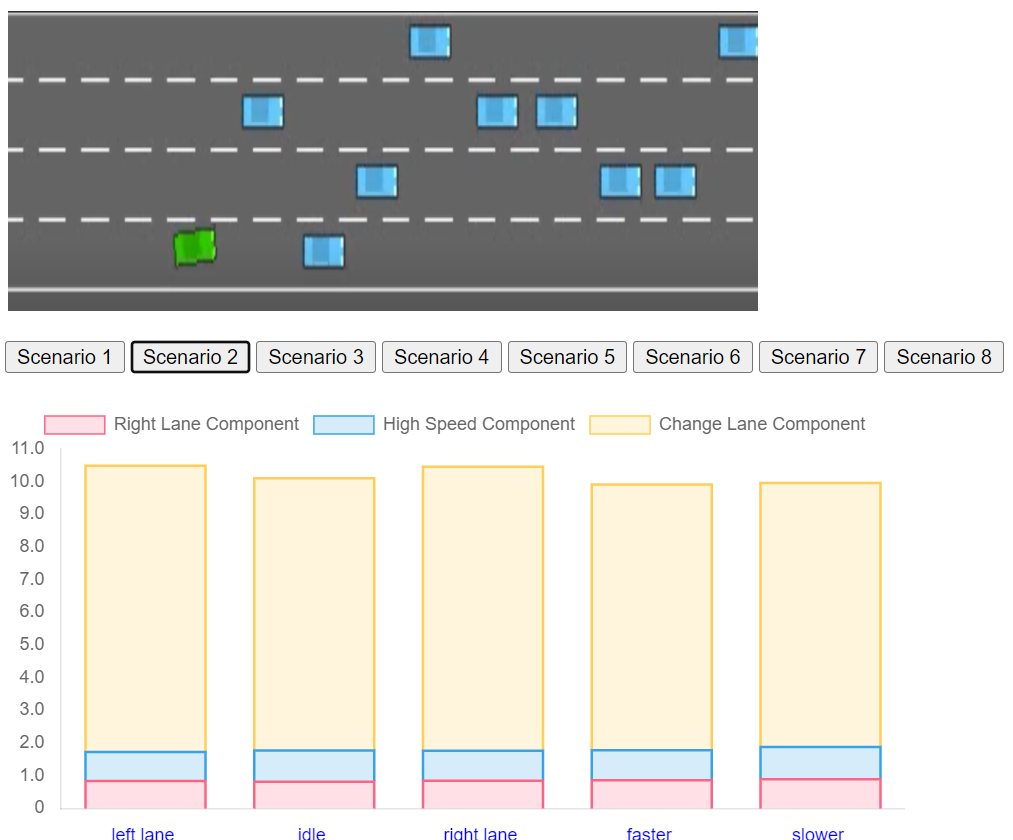}
\caption{A screenshot from the experiment that used the Highway environment. The upper part of the image shows a specific state (``Scenario 2'') extracted from an agent's behavior. The agent controls the green vehicle. The bottom part shows the reward bars corresponding to the state shown above. For each action (shown on the x-axis) the Q-values of the different reward components (depicted in different colors) are shown (y-axis). In this case, it can be observed that the ``change lane'' component is the largest reward component affecting the behavior of this agent in this state. Users could switch to different states by choosing a scenario from the list. The states (scenarios) were chosen based on the summary method (HIGHLIGHTS or frequency-based). For conditions without local explanation, the reward bars were omitted and each scenario showed a short video. }
\label{fig:example_from_survey}
\end{figure}

We used a multi-lane Highway environment (shown in the top part of Figure \ref{fig:example_from_survey}) for our first experiments. The environment can be modified by setting different variables such as the number of vehicles, vehicle density, rewards, speed range, and more. Our settings can be found in the appendix.

In the environment, the RL agent controls the green vehicle. The objective of the agent is to maximize its reward by navigating a multi-lane highway while driving alongside other (blue) vehicles. Positive rewards can be given for each of the following actions: changing lanes (CL), speeding up (SU), and moving to the right-most lane (RML). Therefore, we decided to set $|C|=3$ for each positive reward type.
We trained the agents as described in Section \ref{sec:background}.
The network input is an array of size 25 (5X5) that represents the state. The input layer is followed by two fully connected hidden layers of length 256. The last of these two layers is connected to three heads. Each head consists of a linear layer and outputs a Q-value vector of length of 5 that contains the following: lane left, idle, lane right, faster, slower. 

We trained the following four RL agents which differ in their policies:
\begin{enumerate}
  \item The Good Citizen -  Highest reward for being in the right lane, next to change lane, and lastly to speed up.
    \item Fast And Furious - Highest reward for speeding up, then changing lanes, and lastly to be in the right-most lane. 
    \item Dazed and Confused - Highest reward for changing lanes, next to be in the right-most lane, and lastly to speed up.
    \item Basic - Reward for being in the right-most lane.

  \end{enumerate}

Common to all agents, when crashing a negative reward of -3 is given, and no future rewards can be obtained due to ending the episode. We tried different rewards to get qualitatively different behavior of the agents.
The precise settings of the rewards for the different agents that we used are summarized in Table \ref{tab:reward setting}.
Each agent was trained for 2,000 episodes and each episode consists of 80 steps (or fewer if the agent crashed). 

Our implementation is based on two open source repositories: the Highway environment and an implementation of double DQN.\footnote{\url{https://github.com/eleurent/highway-env}, \url{https://github.com/eleurent/rl-agents}}

\begin{table}[t]
    \centering
    \begin{tabular}{c|c c c c }
        \hline
          & CL  & SU & RML  \\
           & reward & reward& reward \\
        \hline
        The Good Citizen & 3 & 1 & 8  \\
        
         Fast and Furious & 5 & 8 & 1  \\
         
Dazed and Confuse & 8 & 1 & 5 \\
        
         Basic & 0 & 0 & 15 \\
        \hline
        
    \end{tabular}
    \caption{Highway environment - Settings of the four agents.}
    \label{tab:reward setting}
        \vspace{-0.3cm}

\end{table}

The state importance definition for HIGHLIGHTS in the Highway environment is: $I(s) = \max\limits_{a} Q^{\pi}_{(s,a)}- \min\limits_{a}Q^{\pi}_{(s,a)}$, as used in the original HIGHLIGHTS implementation~\cite{amir2018highlights}. 
According to this formulation, a state is considered important if there is a large gap between the expected outcome of the best and worst action available to the agent in the state.
To extract the policy summaries, we ran 2,000 simulations of the trained agent and saved the traces. Summary states are extracted from these traces. 

\subsubsection{Pacman Environment}

In the second experiment, we used the Atari 2600 game MsPacman (Pacman for simplicity) contained in the Arcade Learning Environment \cite{bellemare2013ALE}.

\begin{figure}[t]
\centering
\includegraphics[width=0.7\linewidth]{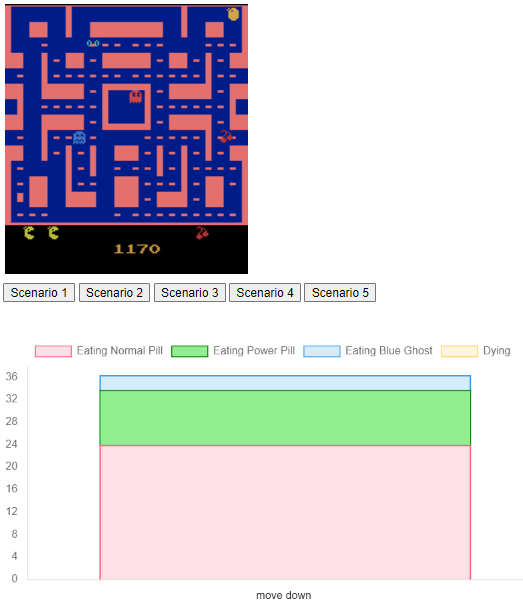}
\caption{A screenshot from the experiment that used the Pacman environment. The upper part of the image shows a specific state extracted from an agent's behavior. The bottom part shows the reward bars corresponding to the state shown above. For the action with the highest Q-value the Q-values of the different reward components (depicted in different colors) are shown (y-axis). In this case, it can be observed that the ``eating normal pill'' component is the largest reward component affecting the behavior of this agent in this state. Users could switch to different states by choosing a scenario from the list. The states (scenarios) were chosen based on the summary method (HIGHLIGHTS or frequency-based). For conditions without local explanation, the reward bars were omitted and each scenario showed a short video. }
\label{fig:Survey_example_Pacman}
    \vspace{-0.2cm}
\end{figure}

For training the agents we build upon the Double DQN implementation in OpenAI baselines \cite{baselines}.
The network architecture is the same as used by \citet{mnih2015human}, which consists of 3 convolutional and 2 fully connected layers, and uses pre-processed pixel values as input.
We implemented reward decomposition (HRA) by sharing the convolutional layers but training individual fully connected layers for each reward component.\footnote{Our implementation
 can be found online: \url{https://github.com/hcmlab/baselines/tree/reward_decomposition.}}

In the game, Pacman has to traverse a labyrinth while avoiding ghosts (see top of Figure~\ref{fig:Survey_example_Pacman}).
Based on the rules of the game, we used four different reward components ($|C|=4$) for the RL agent controlling Pacman: the agent receives a reward of 1 for eating normal pills (NP) and a reward of 5 for eating Power Pills (PP).
Additionally, after eating a PP, the ghosts turn blue and Pacman can eat them. 
The agent receives a reward of 20, 40, 80, 160 for each blue ghost (BG) it eats successively.
Finally, the agent receives a reward of -10 for dying.

We trained three different Pacman agents with the following preferences:
\begin{enumerate}
\item Normal Pill Agent - Highest preference for eating normal pills, next eating power pills and lastly eating blue ghosts.
\item Power Pill Agent - Highest preference for eating power pills, next eating normal pills and eating blue ghost has the same preference.
\item Blue Ghost Agent -  Highest preference for eating blue ghosts, next eating normal pills,  lastly eating power pills.
\end{enumerate}
Each agent was trained for 5 million steps.
To get agents with distinct strategies, we used different weights for the reward components (see Table \ref{tab:reward setting pacman}).
In the Pacman environment, the values of the individual rewards do not directly correlate to the agents' preferences.
For example, the labyrinth contains a huge amount of normal pills compared to power pills and ghosts. 
Therefore, the agent with no specific reward component weights focuses very strongly on normal pills even though the reward value for individual normal pills is the lowest.
To determine what the agents preferred, we observed the Q-values and actions of each agent for several full games before running the experiment.

\begin{table}[t]
    \centering
    \begin{tabular}{c|c c c c }
        \hline
          & Eating NP  & Eating PP & Eating BG & Dying  \\
           & weights & weights & weights \\
        \hline
        NP Agent & 1 & 1 & 1 & 1  \\
         
         PP Agent & 0.01 & 1 & 0.01 & 0.01 \\
        
        BG Agent & 0.1 & 0.1 & 10 & 0.01  \\
        \hline
        
    \end{tabular}
    \caption{How each of the reward components was weighted for our Pacman agents.}
    \label{tab:reward setting pacman}
    \vspace{-0.2cm}
\end{table}

We used HIGHLIGHTS-DIV to generate summaries for the Pacman agents. 
Compared to the basic HIGHLIGHTS algorithm, this approach additionally utilizes a similarity metric (Euclidean distance in our case) to avoid very similar states within the summaries \cite{amir2018highlights}.
Following \citet{huber2020local}, we calculated the importance as $I(s) = \max\limits_{a}Q^{\pi}_{(s,a)}- \operatornamewithlimits{second-highest}\limits_{a}Q^{\pi}_{(s,a)}$.
According to this formulation, a state is considered important if there is a large gap between the expected outcome of the best and second best action available to the agent in the state.
This was done since the game contains several redundant actions. 
Pacman has 9 possible actions - four actions for moving to each side, four for moving diagonally, and one for staying idle. 
However, moving diagonally is translated to moving to a specific side within the game according to an in-game heuristic.  
These redundant actions are often completely ignored by the agents and always have low Q-values.
To extract policy summaries, we ran the trained agent for another 1,000 episodes after the training and recorded the traces.

\subsection{Study Design}
\emph{Experimental conditions}. We conducted two user studies to evaluate the impact of combining global and local explanations, as well as the effect of each method individually. We note that since local explanations are given for specific states, there must be some choice of which states to show the information for. Hence, as a baseline  approach, rather than using HIGHLIGHTS to select states, we used frequency sampling to generate summaries that chooses states for the summary by uniformly sampling from the traces of the agent, as used in the study by~\citet{huber2020local}. Since each state has the same probability of being chosen, in practice states that appear more frequently are more likely to appear in the summary.
Therefore, this is equivalent to selecting states based on the likelihood of their appearance.

We assigned participants randomly to one of four different conditions (summarized in Table \ref{tab:study conditions}):

\begin{itemize}
\item HIGHLIGHTS Summaries (H): In this condition, participants were shown summary videos that were generated by the HIGHLIGHTS algorithm.
We used a context window of 10 states that were shown before and after the chosen state and an interval size of 10 states to prevent directly successive states in the summary.

\item Frequency sampling summaries (FS): 
This condition contained videos similar to condition H, but the states in the middle of the trajectories were chosen based on frequency sampling. 
Moreover, to ensure that the summary is not particularly good or particularly bad we created 10 different summaries of this form for the Highways environment and 5 for Pacman.

\item HIGHLIGHTS + reward decomposition (H+RD): 
Since interpreting reward decomposition takes some time, we did not show videos in this condition.
Instead, participants were only shown the most ``important'' state of each trajectory.
This means that they did not get the context to that state as the video summaries provide.
However, the chosen states were the same states that appeared in the middle of the videos in condition H.
Each chosen state was shown using an image alongside a  bar plot that represents the Q-values of the different reward components. 
In the Highway environment, the bar plot was shown for each available action in the chosen state, as shown in Figure \ref{fig:example_from_survey}. 
Since Pacman contains ambiguous actions, we only showed the bar plot for the agents' chosen action in this environment (see Figure \ref{fig:Survey_example_Pacman}).

\item Frequency sampling summaries + reward decomposition (FS+RD):
This condition was the same as condition H+RD but the shown states were the same states that were uniformly sampled for the middle of the trajectories in the FS condition.

\end{itemize}

\begin{table}[t]
    \centering
    \begin{tabular}{c|cc}
    \hline
         & FS summaries & HIGHLIGHTS \\
        \hline
        No RD & FS & H \\
        RD & FS+RD  & H+RD   \\
        \hline
    \end{tabular}
    \caption{The four study conditions.}
    \label{tab:study conditions}
\end{table}

Following \citet{huber2020local}, we set the size of the summaries for Pacman to $k=5$.
For the Highway environment, we used $k=8$.
Therefore, all participants were shown a summary of the agent's behavior that is composed of 5 or 8 different videos or images regarding the specific agent depending on the environment.

\emph{Procedure.} At first, participants were given an explanation regarding the environment (Highway or Pacman). 
Second, they were given a brief explanation about reinforcement learning and specifically about q-values (the explanation was given in layperson vocabulary\footnotemark \footnotetext{The exact explanations can be seen in the appendix \ref{ap:survey_information}.}).
Lastly, depending on the condition, participants were given information about the type of explanation they will see and an example explanation.
At the end of each instructions phase, the participants were asked to complete a quiz and were only allowed to proceed after answering all questions correctly.
Participants were compensated as follows: they received a \$3 base payment and an additional bonus of 10 cents for each correct answer in the Highway environment and a 30 cent bonus for identifying the preferences of each of the agents correctly in the Pacman environment.

\emph{Task.} 
Participants' task was to assess the preferences of the different agents.
To avoid learning effects, the ordering of the agents was randomized. 
Specifically, participants were asked to rank which of each pair of reward components (e.g., high speed vs. driving in the right lane in the Highway environment or  eating power pills vs. eating normal pills in the Pacman environment) the agent prioritizes or whether it is indifferent between the two options.
If participants have a correct mental model of the agents' strategy, they should be able to rank the different reward components according to the agents' priorities.

Participants were then asked to rate their confidence in each of their answers on a Likert scale from 1 (``not confident at all'') to 5 (``very confident'') and to describe their reasoning in a free-text response.
Lastly, participants rated their agreement on a 7-point Likert scale with the following items adapted from the explanation satisfaction questionnaire proposed by \cite{hoffman2018metrics}:
\begin{enumerate}
\item The videos$\backslash$graphs helped me recognize agent strategies
\item The videos$\backslash$graphs contain sufficient detail for recognizing agent strategies
\item The videos$\backslash$graphs contain irrelevant details 
\item The videos$\backslash$graphs were useful for the tasks I had to do in this survey
\item The specific scenarios shown in the videos$\backslash$images were useful for the tasks I had to do in this survey.
\end{enumerate}

\emph{Participants}. We recruited participants through Amazon Mechanical Turk (N = 164 for each environment). We excluded participants who did not answer the attention question correctly, as well as participants who completed the survey in less than 7 minutes or in less than two standard deviations from the mean completion time in their condition. These values were based on a pilot study.

After screening, we had 127 and 159 participants in the Highway environment and the Pacman environment respectively (mean age = 36 years for both environments, 58 and 88 female in the Highway environment and the Pacman environment respectively,  all from the US, UK, or Canada).

\section{Results}
\begin{figure}[t]
\begin{minipage}{0.49\linewidth}
\centering
\includegraphics[width=\linewidth]{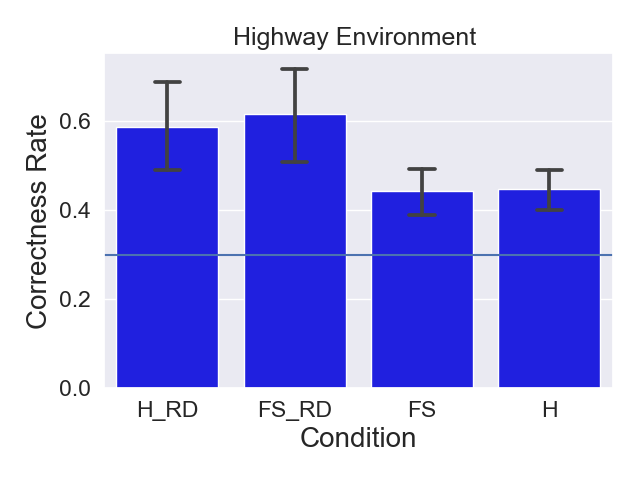}
(A)
\end{minipage}
\begin{minipage}{0.49\linewidth}
\centering
\includegraphics[width=\linewidth]{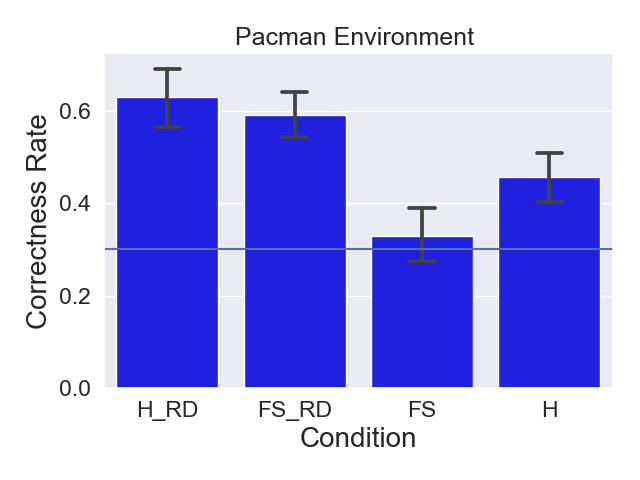}
(B)
\end{minipage}
\caption{Participants' mean success rate in identifying the preferences averaged over all agents by condition in the Highway environment (A) and in the Pacman environment (B). The error bars show the 95\% CI. }
\label{fig:results}
    \vspace{-0.2cm}
\end{figure}

To measure participants' ability to asses the agents' preferences, we calculated the mean fraction of correct reward component comparisons, i.e., their correctness rate, for each condition (see Figure~\ref{fig:results}).
We tested our hypotheses using the non-parametric, one-sided Mann-Whitney $U$ test.  Only when comparing the individual explanation conditions H and FS+RD, we used a two-sided test since we did not have an hypothesis as to which method will be better.

We found that reward decomposition improved participants' ability to asses the agents' preferences in both environments.
In the Highway environment, the combination of FS+RD 
led to significantly improved performance compared to FS 
(FS+RD vs. FS: $U$=$736$, $p$=$0.014$; FS+RD vs. H: $U$=$791$, $p$=$0.017$). 
Similarly, H+RD 
led to significantly improved performance compared to FS and H ($U$=$532$, $p$=$0.014$ and $U$=$571$, $p$=$0.018$, respectively).
Similarly, in the Pacman environment, the combination of FS+RD 
significantly improved participants' performance compared to FS or H
($U$=$1187$ and $U$=$1176$ respectively, $p$$<$$0.001$ for both) and participants in the H+RD condition 
performed better compared to FS and H ($U$=$1286$ and $U$=$1332$ respectively, $p$$<$$0.001$ for both).
Some of the participants explicitly referred to reward decomposition as being helpful for the task, e.g., ``In each of the scenarios the graph clearly shows the preference for normal pills followed by eating power pills. Eating ghosts was a minor section on the graph.''

The HIGHLIGHTS summaries only contributed to participants' mental model of the agents in the Pacman environment. 
Here, there was a significant difference between condition H and condition FS ($U$=$935$, $p$=$0.002$). In some explanations given by participants it seemed that the HIGHLIGHTS summary displayed information that was useful for inferring preference, e.g., one participant wrote ``the pacman would go for a power pill, eat it and turn around'' when explaining their answers about the Power Pill agent preferences.
In the Highway environment we did not observe a significant difference between those two conditions.

In both environments, the combined explanation did not outperform reward decomposition alone.
There were no significant differences between H+RD and FS+RD in either of the environments when aggregated across all agents.
However, in the Highway environment, our results indicate that the combination of H+RD helped asses the agent's preferences when the difference between the reward types was minor. 
For example, when assessing the agent ``Fast and Furious'' that was trained according to the rewards of 8 points for speed up vs. 5 points for changing lanes, participants who were shown H+RD succeeded about half of the times ($M$=$0.55$, $95\%$ $CI$=$(0.41,0.69)$) compared to participants in conditions FS+RD, FS or H that had a lower success rate ($M$=$0.44$, $95\%$ $CI$=$(0.34,0.54)$; $M$=$0.18$, $95\%$ $CI$=$(0.03,0.33)$; $M$=$0.26$, $95\%$ $CI$=$(0.1,0.42)$, respectively). 
Similarly, for the BG agent in Pacman, there was only a small difference between the Q-values for blue ghost and normal pill. 
Participants in conditions H ($M$=$0.3$, $95\%$ $CI$=$(0.2,0.5)$) and H+RD ($M$=$0.31$, $95\%$ $CI$=$(0.17,0.46)$) were better at correctly identifying the blue ghost as more important than the participants in conditions FS+RD ($M$=$0.2$, $95\%$ $CI$=$(0.17,0.46)$)  and FS ($M$=$0.12$, $95\%$ $CI$=$(0,0.23)$).
This indicates that even though our overall results do not show that the combination of H+RD is significantly better, there were cases in which this combination helped.

In the Highway environment, participants' confidence and satisfaction ratings were above the neutral rating ($>3$) but there was no difference between the conditions.
In the Pacman environment, the confidence and satisfaction values of participants were also above neutral as seen in Figure~\ref{fig:satisfaction} in Appendix \ref{ap:Satisfaction}.
However, here the explanation conditions FS+RD, H+RD, and H had higher mean ratings ($M$ between $5.34$ and $5.39$) than the baseline condition (condition FS with $M$=$4.93$).
In general, while participants' objective performance was better with RD compared to video-based policy summaries, this did not lead to an increase in the subjective measures.

\section{Discussion and Future Work}
This paper presented a new approach for describing the behavior of RL agents, which integrates HIGHLIGHTS, a global policy summary with reward decomposition. We conducted user studies in two environments to evaluate the contribution of this approach to people's ability to analyze agent preferences. Our results show that reward decomposition was particularly helpful for this task and that HIGHLIGHTS also led to improvement in participants' performance, but only in certain situations. 

In previous work, HIGHLIGHTS summaries were integrated with saliency maps, but a user study showed that saliency maps did not provide much benefit to users' understanding of agent behavior~\cite{huber2020local}. We hypothesized that reward decomposition may be more beneficial for several reasons. First, saliency maps describe what features of the state the agent pays attention to, but it is often hard to infer how this information affects the agent's decisions, especially for laypeople. Reward decomposition has the benefit of explicitly describing what values the agent expects to get in a way that reflects its preferences for different reward components. Moreover, saliency maps are a post-hoc method and may not be faithful to the underlying model~\cite{rudin19,huber2022benchmarking} while reward decomposition values are learned through the agent's training and reflect its true decision-making policy. 
Another difference between our integration of global and local information and the one used in the study by~\citet{huber2020local} is the use of static images rather than videos. We chose this approach based on the findings from their study which identified the use of videos as one possible limitation, as the local information is harder to discern when looking at dynamic videos. These differences between the two local explanation methods may explain the discrepancy between the results in the previous study, which showed little contribution of the local explanation, and our study, where local explanations in the form of reward decomposition contributed substantially to participants' understanding. 

Our studies showed a fairly limited contribution of HIGHLIGHTS summaries to participants' performance compared to prior works~\cite{amir2018highlights,huber2020local}, with improved performance observed only in some scenarios in the Pacman environment. A possible explanation for the limited contribution of HIGHLIGHTS is that reward decomposition may have been too suited for our experimental task. Since reward decomposition was already highly effective in conveying agent preferences, the selection of states for the summary was less important. We further hypothesize that HIGHLIGHTS did not improve performance over frequency-based summaries in the Highway environment since the environment itself is somewhat limited in terms of different agent behaviors.

The different explanation methods were found to have different contributions depending on the context, that is, the environment and the task. Future work could explore how the characteristics of the environment affect the usefulness of different explanation methods. For instance, the usefulness of policy summaries may depend on the complexity of behaviors that agents may deploy in a domain, while the usefulness of reward decomposition may depend on the extent to which the reward function can be decomposed into reward components that are meaningful to users. It would also be interesting to design methods to elicit reward components from users and formalize them, such that the reward function can be decomposed in a way that fits users' mental models of the environment. In addition, future work could explore other combinations of global and local explanation methods, potentially going beyond the integration of two methods at a time and providing users with a suite of explanation methods. These can then be presented depending on the context and the users' goals. Another notable finding is that the use of different explanation methods did not result in substantial differences in the subjective measures, i.e., the confidence and satisfaction ratings. This finding emphasizes the importance of using objective performance measures, while also raising questions for future work such as how to provide feedback to users regarding their mental models of agent behavior.  

\section{Acknowledgements}
We thank Julian Stockmann and Simone Pompe for their help with implementing HRA for MsPacman.

\bibliography{aaai23} instead or the References section will not appear in your paper

\begin{thebibliography}{21}
\providecommand{\natexlab}[1]{#1}

\bibitem[{Amir and Amir(2018)}]{amir2018highlights}
Amir, D.; and Amir, O. 2018.
\newblock Highlights: Summarizing agent behavior to people.
\newblock In \emph{Proceedings of the 17th International Conference on
  Autonomous Agents and MultiAgent Systems}, 1168--1176.

\bibitem[{Amir, Doshi-Velez, and Sarne(2019)}]{amir2019summarizing}
Amir, O.; Doshi-Velez, F.; and Sarne, D. 2019.
\newblock Summarizing agent strategies.
\newblock \emph{Autonomous Agents and Multi-Agent Systems}, 33(5): 628--644.

\bibitem[{Anderson et~al.(2019)Anderson, Dodge, Sadarangani, Juozapaitis,
  Newman, Irvine, Chattopadhyay, Fern, and Burnett}]{anderson2019explaining}
Anderson, A.; Dodge, J.; Sadarangani, A.; Juozapaitis, Z.; Newman, E.; Irvine,
  J.; Chattopadhyay, S.; Fern, A.; and Burnett, M. 2019.
\newblock Explaining reinforcement learning to mere mortals: An empirical
  study.
\newblock \emph{arXiv preprint arXiv:1903.09708}.

\bibitem[{Bellemare et~al.(2013)Bellemare, Naddaf, Veness, and
  Bowling}]{bellemare2013ALE}
Bellemare, M.~G.; Naddaf, Y.; Veness, J.; and Bowling, M. 2013.
\newblock The Arcade Learning Environment: An Evaluation Platform for General
  Agents.
\newblock \emph{J. Artif. Intell. Res.}, 47: 253--279.

\bibitem[{Booth, Muise, and Shah(2019)}]{booth2019evaluating}
Booth, S.; Muise, C.; and Shah, J. 2019.
\newblock Evaluating the Interpretability of the Knowledge Compilation Map:
  Communicating Logical Statements Effectively.
\newblock In \emph{IJCAI}, 5801--5807.

\bibitem[{Dhariwal et~al.(2017)Dhariwal, Hesse, Klimov, Nichol, Plappert,
  Radford, Schulman, Sidor, Wu, and Zhokhov}]{baselines}
Dhariwal, P.; Hesse, C.; Klimov, O.; Nichol, A.; Plappert, M.; Radford, A.;
  Schulman, J.; Sidor, S.; Wu, Y.; and Zhokhov, P. 2017.
\newblock OpenAI Baselines.
\newblock \url{https://github.com/openai/baselines}.

\bibitem[{Hilton et~al.(2020)Hilton, Cammarata, Carter, Goh, and
  Olah}]{hilton2020understanding}
Hilton, J.; Cammarata, N.; Carter, S.; Goh, G.; and Olah, C. 2020.
\newblock Understanding RL Vision.
\newblock \emph{Distill}, 5(11): e29.

\bibitem[{Hoffman et~al.(2018)Hoffman, Mueller, Klein, and
  Litman}]{hoffman2018metrics}
Hoffman, R.~R.; Mueller, S.~T.; Klein, G.; and Litman, J. 2018.
\newblock Metrics for explainable AI: Challenges and prospects.
\newblock \emph{arXiv preprint arXiv:1812.04608}.

\bibitem[{Huang et~al.(2019)Huang, Held, Abbeel, and
  Dragan}]{huang2019enabling}
Huang, S.~H.; Held, D.; Abbeel, P.; and Dragan, A.~D. 2019.
\newblock Enabling robots to communicate their objectives.
\newblock \emph{Autonomous Robots}, 43(2): 309--326.

\bibitem[{Huber, Limmer, and André(2022)}]{huber2022benchmarking}
Huber, T.; Limmer, B.; and André, E. 2022.
\newblock Benchmarking Perturbation-Based Saliency Maps for Explaining Atari
  Agents.
\newblock \emph{Frontiers in Artificial Intelligence}, 5.

\bibitem[{Huber, Schiller, and Andr{\'e}(2019)}]{huber2019}
Huber, T.; Schiller, D.; and Andr{\'e}, E. 2019.
\newblock Enhancing Explainability of Deep Reinforcement Learning Through
  Selective Layer-Wise Relevance Propagation.
\newblock In Benzm{\"u}ller, C.; and Stuckenschmidt, H., eds., \emph{KI 2019:
  Advances in Artificial Intelligence}, 188--202. Cham: Springer International
  Publishing.
\newblock ISBN 978-3-030-30179-8.

\bibitem[{Huber et~al.(2021)Huber, Weitz, Andr{\'{e}}, and
  Amir}]{huber2020local}
Huber, T.; Weitz, K.; Andr{\'{e}}, E.; and Amir, O. 2021.
\newblock Local and global explanations of agent behavior: Integrating strategy
  summaries with saliency maps.
\newblock \emph{Artif. Intell.}, 301: 103571.

\bibitem[{Juozapaitis et~al.(2019)Juozapaitis, Koul, Fern, Erwig, and
  Doshi-Velez}]{juozapaitis2019explainable}
Juozapaitis, Z.; Koul, A.; Fern, A.; Erwig, M.; and Doshi-Velez, F. 2019.
\newblock Explainable reinforcement learning via reward decomposition.
\newblock In \emph{IJCAI/ECAI Workshop on Explainable Artificial Intelligence}.

\bibitem[{Lage et~al.(2019)Lage, Lifschitz, Doshi-Velez, and
  Amir}]{lage2019exploring}
Lage, I.; Lifschitz, D.; Doshi-Velez, F.; and Amir, O. 2019.
\newblock Exploring computational user models for agent policy summarization.
\newblock In \emph{IJCAI: proceedings of the conference}, volume~28, 1401. NIH
  Public Access.

\bibitem[{Madumal et~al.(2020)Madumal, Miller, Sonenberg, and
  Vetere}]{madumal2020distal}
Madumal, P.; Miller, T.; Sonenberg, L.; and Vetere, F. 2020.
\newblock Distal Explanations for Model-free Explainable Reinforcement
  Learning.
\newblock \emph{arXiv preprint arXiv:2001.10284}.

\bibitem[{Mnih et~al.(2015)Mnih, Kavukcuoglu, Silver, Rusu, Veness, Bellemare,
  Graves, Riedmiller, Fidjeland, Ostrovski et~al.}]{mnih2015human}
Mnih, V.; Kavukcuoglu, K.; Silver, D.; Rusu, A.~A.; Veness, J.; Bellemare,
  M.~G.; Graves, A.; Riedmiller, M.; Fidjeland, A.~K.; Ostrovski, G.; et~al.
  2015.
\newblock Human-level control through deep reinforcement learning.
\newblock \emph{nature}, 518(7540): 529--533.

\bibitem[{Molnar(2022)}]{molnar2022}
Molnar, C. 2022.
\newblock \emph{Interpretable Machine Learning}.
\newblock 2 edition.

\bibitem[{Puri et~al.(2020)Puri, Verma, Gupta, Kayastha, Deshmukh,
  Krishnamurthy, and Singh}]{puri2020}
Puri, N.; Verma, S.; Gupta, P.; Kayastha, D.; Deshmukh, S.; Krishnamurthy, B.;
  and Singh, S. 2020.
\newblock Explain Your Move: Understanding Agent Actions Using Specific and
  Relevant Feature Attribution.
\newblock In \emph{8th International Conference on Learning Representations,
  {ICLR}}. OpenReview.net.

\bibitem[{Rudin(2019)}]{rudin19}
Rudin, C. 2019.
\newblock Stop explaining black box machine learning models for high stakes
  decisions and use interpretable models instead.
\newblock \emph{Nat. Mach. Intell.}, 1(5): 206--215.

\bibitem[{Van~Hasselt, Guez, and Silver(2016)}]{van2016deep}
Van~Hasselt, H.; Guez, A.; and Silver, D. 2016.
\newblock Deep reinforcement learning with double q-learning.
\newblock In \emph{Proceedings of the AAAI conference on artificial
  intelligence}, 2094--2100.

\bibitem[{Van~Seijen et~al.(2017)Van~Seijen, Fatemi, Romoff, Laroche, Barnes,
  and Tsang}]{van2017hybrid}
Van~Seijen, H.; Fatemi, M.; Romoff, J.; Laroche, R.; Barnes, T.; and Tsang, J.
  2017.
\newblock Hybrid reward architecture for reinforcement learning.
\newblock \emph{arXiv preprint arXiv:1706.04208}.

\end{thebibliography}

\clearpage

\appendix

\section{Sanity Check}
\label{ap:sanity_checks}
We first wanted to ensure that the changes made to the neural network architecture to incorporate reward decomposition resulted in learning comparable to that of an agent trained without decomposed rewards. To this end, in the Highway environment, we compared the cumulative rewards of two RL agents that differ only by their neural network architecture. To correctly compare the single-value reward to the decomposed reward, we normalized the decomposed rewards so that the sum of all available reward components approximately equals the available single-value reward. Table ~\ref{tab:valus_Original_vs_Multi} presents the main parameters that were used for each RL agent.
The results show that the average reward of an RL agent with multiple heads is in the same range as an RL agent that uses a traditional neural network (see Table~\ref{tab:valus_Original_vs_Multi}).
We can conclude from this check that in our domain using reward decomposition did not harm the performance of the agent.

\section {Satisfaction Results}
\label{ap:Satisfaction}

\begin{figure}[h]
\begin{minipage}{0.49\linewidth}
\centering

\includegraphics[width=\linewidth]{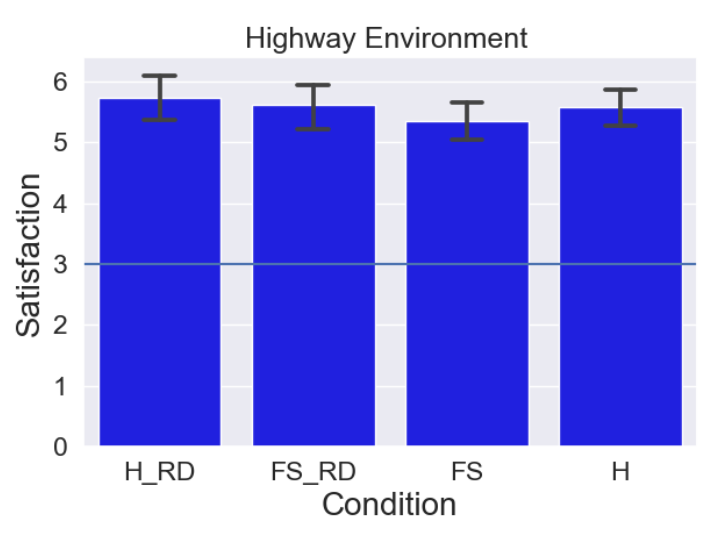}

(A)

\end{minipage}
\begin{minipage}{0.49\linewidth}
\centering

\includegraphics[width=\linewidth]{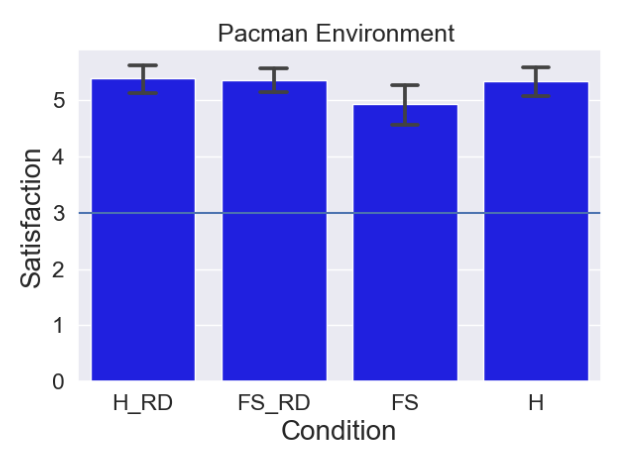}

(B)
\end{minipage}
\caption{Participants' explanation satisfaction by condition in the Highway environment (A) and in the Pacman environment (B). The error bars show the 95\% CI. }
\label{fig:satisfaction}
    \vspace{-0.2cm}
\end{figure}

The results of the Explanation Satisfaction scale are shown in Figure \ref{fig:satisfaction}. 

\section{Survey Information}
\label{ap:survey_information}
The following figures are screenshots from the Pacman environment survey under the condition of Highlight explanation.
As an example, we show the survey only for the Highlight condition and only included one of the agents.
The participants were shown multiple different agents and depending on their condition they saw explanations as shown in Figure \ref{fig:Survey_example_Pacman}.
The Highway environment survey was conducted similarly to the survey that is presented here.

\begin{figure}[]
\centering
\includegraphics[width=0.7\linewidth]{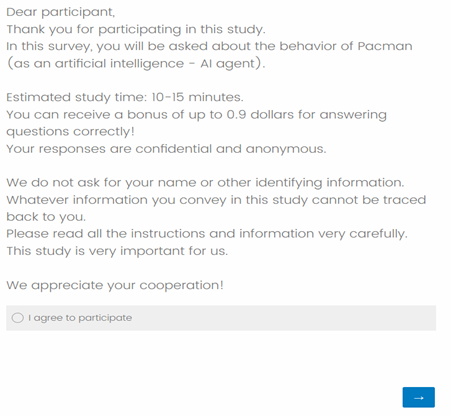}
\end{figure}
\begin{figure}[t]
\centering
\includegraphics[width=0.7\linewidth]{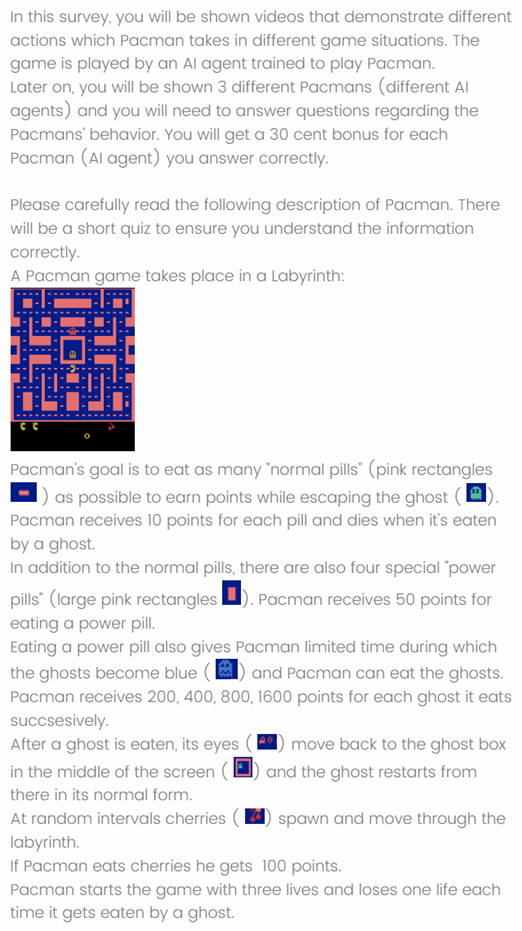}
\end{figure}
\begin{figure}[t]
\centering
\includegraphics[width=0.7\linewidth]{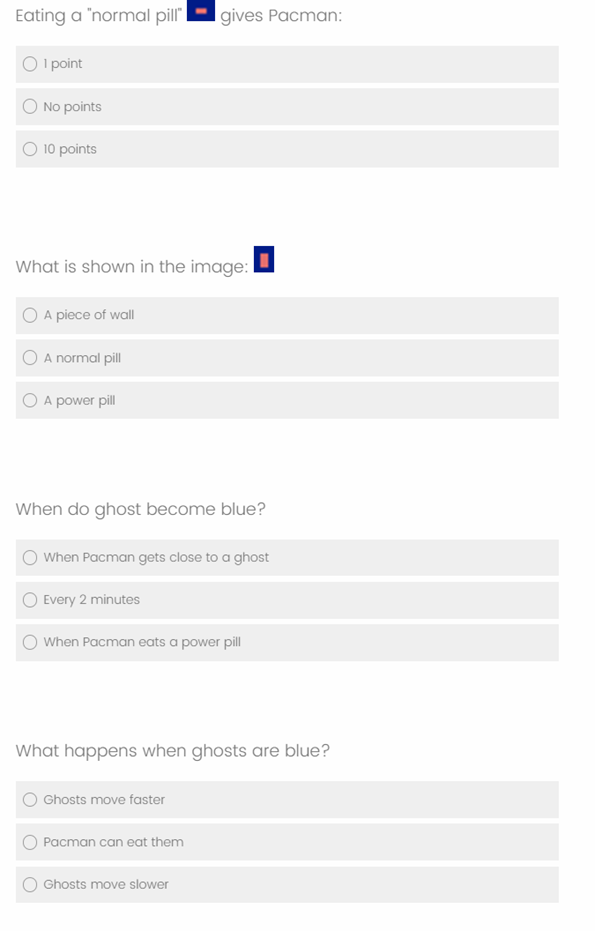}
\end{figure}
\begin{figure}[t]
\centering
\includegraphics[width=0.7\linewidth]{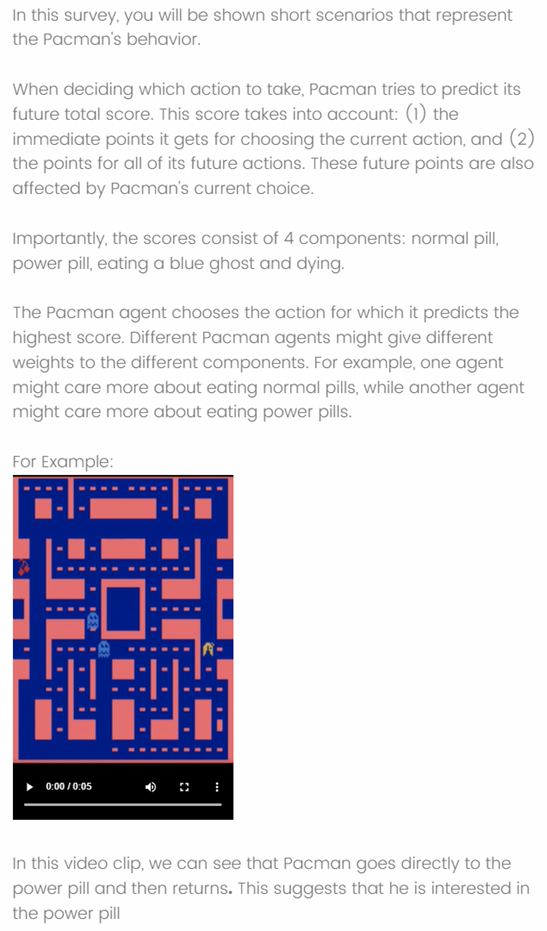}
\end{figure}
\begin{figure}[t]
\centering
\includegraphics[width=0.7\linewidth]{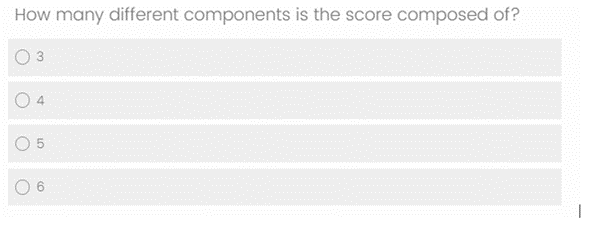}
\end{figure}
\begin{figure}[t]
\centering
\includegraphics[width=0.7\linewidth]{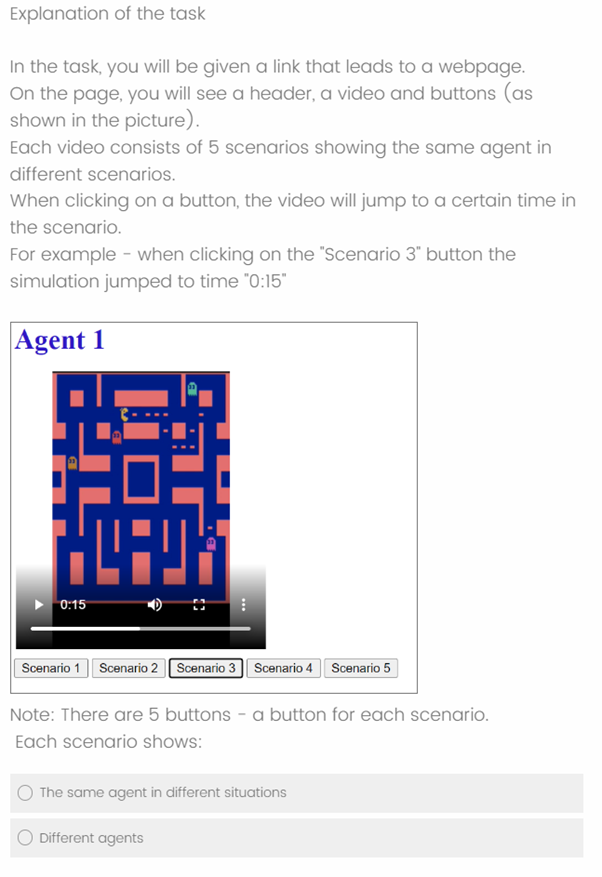}
\end{figure}
\begin{figure}[t]
\centering
\includegraphics[width=0.7\linewidth]{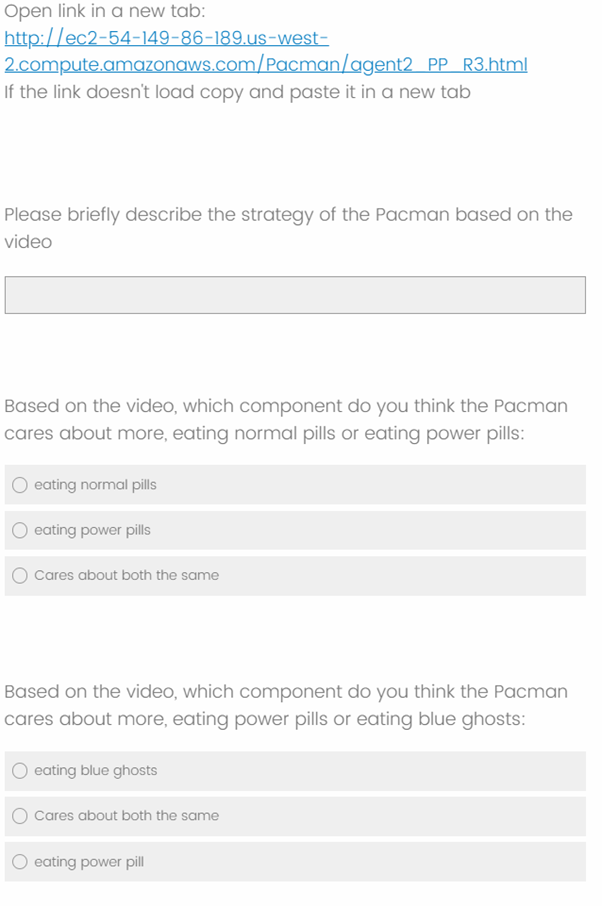}
\end{figure}
\begin{figure}[t]
\centering
\includegraphics[width=0.7\linewidth]{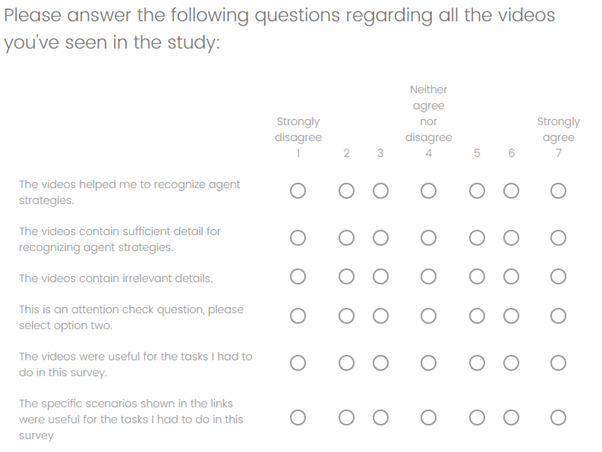}
\end{figure}

\clearpage

\begin{table*}[!ht]
\centering
\begin{tabular}{c|c c }
\hline

 & Original & Multiple Heads \\
\hline
Type & Multi Layer Perceptron & Multi Layer Perceptron\\

Method & Epsilon Greedy & Epsilon Greedy\\

Loss function & L2 & L2\\

Duration of each episode&40 time stamps&40 time stamps\\

Number of lanes & 4 & 4\\

Number of vehicles & 30 & 30\\
\hline
& & head 1 right lane=5\\
& right lane=5&  head 1 high speed=0\\
& &head 1 lane change=0\\ \cline{2-3}
& & head 2 right lane=0\\
Reward& high speed=5&  head 2 high speed=5\\
& & head 2 lane change=0\\ \cline{2-3}
& & head 3 right lane=0\\
& lane change=5 &  head 3 high speed=0\\
& &head 3 lane change=5\\ 
\hline
Reward normalization range &[0,1]&[0,1/3]-for each reward component\\

Number of episodes&2000&2000\\

Average result of reward & 38 & 39\\

\hline
\end{tabular}
\caption{Main values and results of RL agent with original neural network vs. multi head neural network}
\label{tab:valus_Original_vs_Multi}
\end{table*}

\section{Highway Environment}
\label{ap:highway_env}

In the Highway environment we used the following parameters:
\begin{itemize}
\item Number of lanes = 4
\item Vehicles count = 30
\item Duration = 40
\item Ego spacing = 2
\item Vehicles density =1
\item Simulation frequency= 15
\item Policy frequency=15
\end{itemize}
\end{document}